\def\year{2019}\relax
\documentclass[letterpaper]{article} 
\usepackage{aaai19}  
\usepackage{times}  
\usepackage{helvet}  
\usepackage{courier}  
\usepackage{url}  
\usepackage{graphicx}  
\usepackage{multirow}
\usepackage{array}
\usepackage{amssymb}
\usepackage{amsmath}
\usepackage{algorithm}
\usepackage{algorithmic}
\newcommand{\tabincell}[2]{\begin{tabular}{@{}#1@{}}#2\end{tabular}}
\begin{document}
%
\title{Semantic Adversarial Network with Multi-scale Pyramid Attention \\ for Video Classification}
\author{De Xie\textsuperscript{\rm 1}, Cheng Deng\textsuperscript{\rm 1}\thanks{Corresponding author.}, Hao Wang\textsuperscript{\rm 1}, Chao Li\textsuperscript{\rm 1}, Dapeng Tao\textsuperscript{\rm 2}\\
	\textsuperscript{\rm 1}School of Electronic Engineering, Xidian University, Xi'an 710071, China\\
	\textsuperscript{\rm 2}School of Information Science and Engineering, Yunnan University, Kunming 650091, China\\
	\{xiede.xd, chdeng.xd, haowang.xidian,\}@gmail.com, li\_chao@stu.xidian.edu.cn,  dapeng.tao@gmail.com\\
}
\maketitle

\begin{abstract}
Two-stream architecture have shown strong performance in video classification task. The key idea is to learn spatio-temporal features by fusing convolutional networks spatially and temporally. However, there are some problems within such architecture. First, it relies on optical flow to model temporal information, which are often expensive to compute and store. Second, it has limited ability to capture details and local context information for video data. Third, it lacks explicit semantic guidance that greatly decrease the classification performance. In this paper, we proposed a new two-stream based deep framework for video classification to discover spatial and temporal information only from RGB frames, 
moreover, the multi-scale pyramid attention (MPA) layer and the semantic adversarial learning (SAL) module is introduced and integrated in our framework. The MPA enables the network capturing global and local feature to generate a comprehensive representation for video, and the SAL can make this representation gradually approximate to the real video semantics in an adversarial manner. Experimental results on two public benchmarks demonstrate our proposed methods achieves state-of-the-art results on standard video datasets.
\end{abstract}

\section{Introduction}
\label{Introduction}
Video classification is a fundamental task in computer vision community, and it serves as an important basis for high-level tasks, such as video caption \cite{Wang_2018_CVPR}, action detection \cite{rene2017temporal}, and video tracking \cite{Feng_2018_CVPR}. Significant progress on video classification has been made by deep learning on account of the powerful modeling capability of deep convolutional neural networks that obtain superior performance than those hand-crafted representation based methods. However, compared with other visual tasks \cite{Li_2018_CVPR,fan2018compressed,deng2018triplet,yang2018shared}, video classification should consider not only static spatial information in each frame but also dynamic temporal information between frames. Although deep convolutional neural networks can model spatial information well, it is limited ability to capture temporal information only from frame sequence. Therefore, how to model spatial and temporal information effectively with deep learning framework is still a challenging problem.

\begin{figure}[!t]
	\centering
	\includegraphics[width=.95\columnwidth]{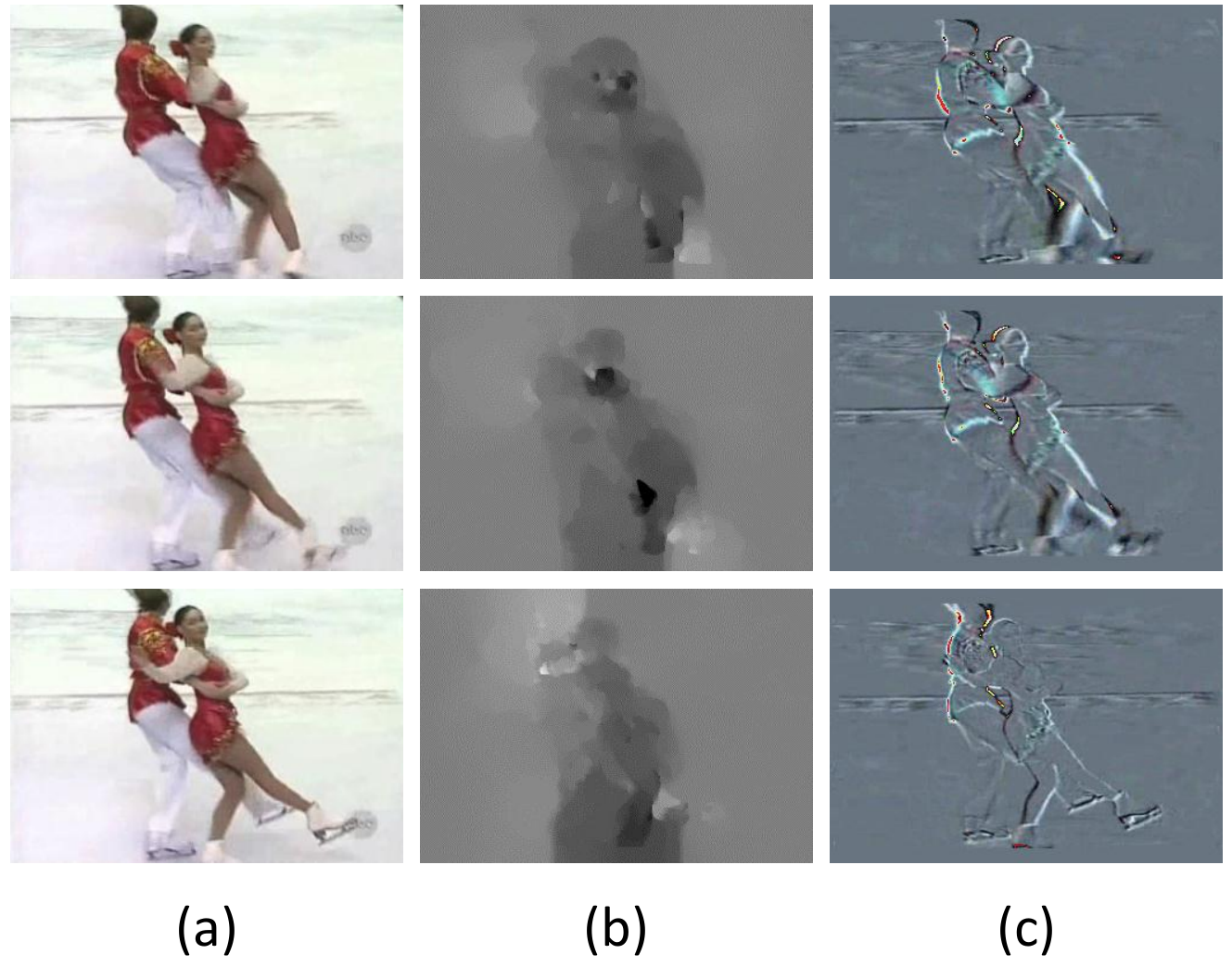}
	\caption{Modeling temporal information with images. (a) input frames; (b) The optical flows between these frames, (c) The differential images between multiple video frames.}
	\label{fig: diff}
\end{figure}

Video classification methods based on deep learning can be divided into three different categories. The first category relies on a combination of multiple input modalities, which models spatial and temporal information, respectively. The two-stream CNN \cite{simonyan2014two} is a groundbreaking work of this category, which captures static spatial information and dynamic temporal information with different streams from multi-modality input, usually RGB images and optical flow. Due to its prominent performance, many state-of-the-art methods can be considered as variants and improvements of this paradigm. However, this method suffers from the heavy reliance on optical flow to model temporal information, which are often expensive to compute and store. To overcome this limitation, the second category takes 2D CNN with temporal models on top such as LSTM \cite{donahue2015long}, temporal convolution \cite{yue2015beyond} and sparse sampling and aggregation \cite{wang2016temporal}. This category usually extracts features from different frames with 2DCNN, then captures the relationship between these features using temporal models. Such type of method more intuitive but lacks capacity to obtain local dynamic information and global context information. The third category is based on 3DCNN \cite{tran2015learning,ji20133d}, which employs 3D convolutions and 3D pooling operations to directly learn spatio-temporal features from stacked RGB volumes. Such methods seem to having ability solve the problem of spatio-temporal modeling but the performance is still worse than two-steam CNN based methods. Meanwhile, 3DCNN based methods also suffer from a large number of parameters and huge computation burden. More important, all three categories methods ignore utilizing the semantic information embodied in video, which leads to limited generalization performance. In fact, RGB frames contain abundant of semantic information, which can greatly improve classification performance. In addition, Inspired by prior work \cite{wang2016temporal}, we find the RGB differential image between multiple video frames have sufficient ability to model temporal information, which is less computational cost than optical flows. As shown in Figure~\ref{fig: diff}, the RGB differential images are sensitive to the part of the motion in the video, which means that the details of RGB differential images have ability to model the temporal information.

In this paper, we propose a new two-stream based architecture to address all mentioned problems. Specially, we design a \textit{Spatial Network} to model spatial information which takes RGB frames as input and a \textit{Temporal Network} to model temporal information which exploits differential images as input. In order to obtain more discriminative representation, we design a multi-scale pyramid attention (MPA) layer to capture multi-scale features from different stage of \textit{Spatial Network} and \textit{Temporal Network}, and then combine these multi-scale information into new representation. In addition, we devise semantic adversarial learning (SAL) module aiming to guide \textit{Spatial Network} and \textit{Temporal Network} to learn more discriminative and semantic video representation. 
Overall, the main contribution of the proposed method can be summarized as follows:

\begin{itemize}
	\item We propose a new deep architecture for video classification, which contains \textit{Spatial Network} and \textit{Temporal Network} only taking RGB frames as input, which significantly reduced computational complexity without sacrificed.	
	\item We devise a multi-scale pyramid attention (MPA) layer that conducts attention-driven multi-scale features extraction and it is pluggable that can be easily embedded to other CNNs based architecture.
	\item We introduce a semantic adversarial learning (SAL) module, which can make fully use of video semantic information and guide video representation learning in adversarial manner.
	\item Experimental results on two public benchmarks for action recognition, HMDB51 and UCF101, highlight the advantages of our method and obtain improved performance compared to state-of-the-art methods.
\end{itemize}
\begin{figure*}[!t]
	\centering
	\includegraphics[width=17cm,height=7.6cm]{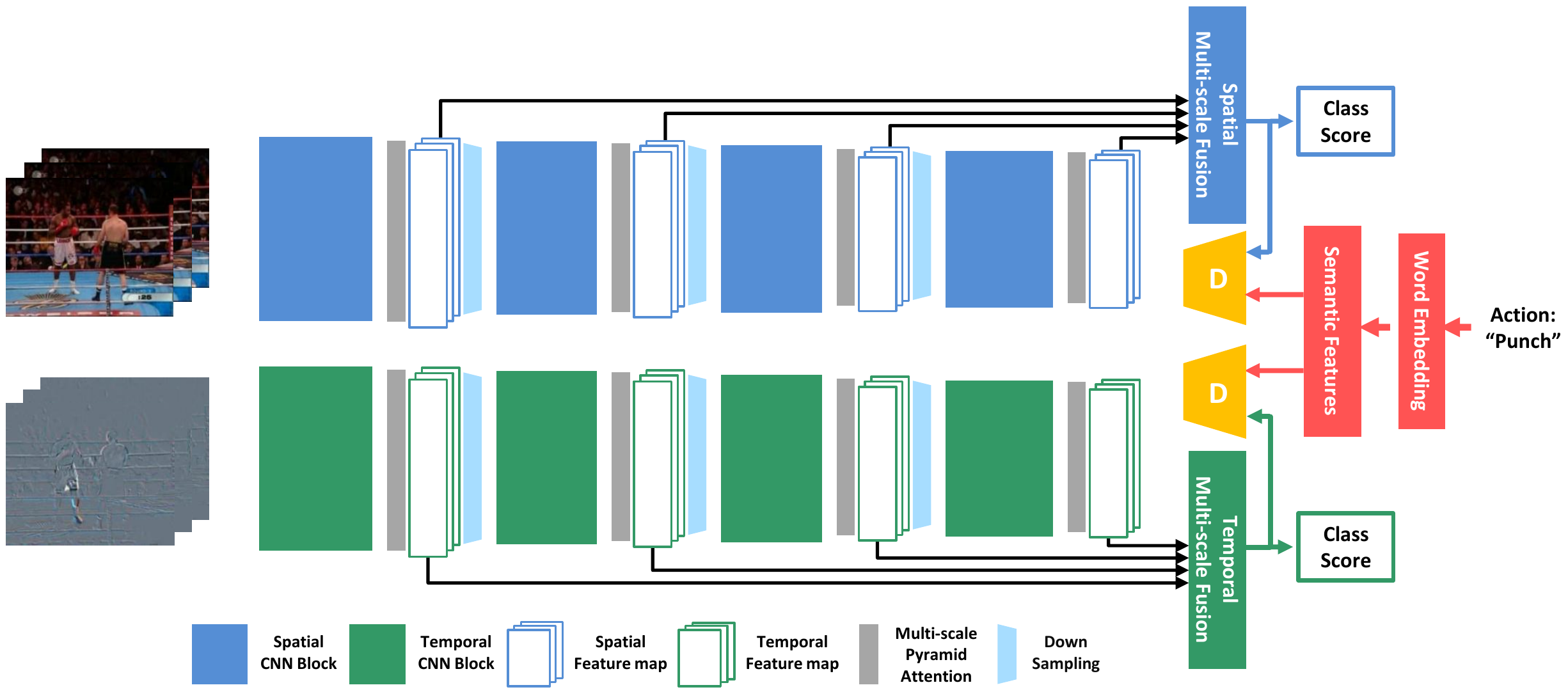}
	\vspace{-0.2cm}
	\caption{Our proposed framework. It can be divided into three components. The top branch is \textit{Spatial Network}, which can model spatial information. The middle branch is \textit{Semantic encoder}, which provides video semantic features. The bottom branch is \textit{Temporal Network}, which can model temporal information. The multi-scale pyramid attention and semantic adversarial learning is introduced in our framework to learn discriminative and semantically rich representation from videos.} 
	\label{fig: framework}
\end{figure*}

\section{Related Work}
\label{sec: RelatedWork}
Video classification has received sustained attention in recent years, and has spawned lots of excellent works\cite{yang2017discriminative,yang2017latent,yang2016multi}. Traditional methods rely on hand-craft visual features such as Motion Boundary Histogram (MBH) \cite{dalal2006human} and improved Dense Trajectory (iDT) \cite{wang2013action} which lack the discriminative capacity to classify complex videos. Deeply learned features is proved more powerful than hand-craft features which can achieve superior performance. 

There are many works have been trying to design effective deep architecture for video classification. For example, Karpathy \textit{et al.} \cite{karpathy2014large} showed the first large-scale experiment on training deep convolutional neural networks from a large video dataset, Sports-1M. Two-Stream \cite{simonyan2014two}, as a significant breakthrough method, containing spatial and temporal nets to model appearance and motion information respectively. Wang \textit{et al.} \cite{wang2016temporal} designed temporal segment network to perform sparse sampling and temporal fusion, which aims to learn from the entire video. Wang \textit{et al.} \cite{wang2017exploring} further improved this architecture by integrating appearance information, short-term and long-term motion information, which achieve outstanding classification performance. However, these methods used optical flows to caption motion which is time consuming. In order to capture the motion information directly from RGB frames, a set of methods have been proposed to use 3DCNN \cite{tran2015learning}, containing 3D convolution filters and 3D pooling layers, to model spatial and temporal information simultaneously. Although it is intuitive, but in fact, spatial information and temporal information may interfere with each other during the modeling process. So, it is still unclear whether this pattern could efficiently model spatial and temporal relation. To explicitly modeling spatial and temporal information, CNN-LSTM \cite{shi2017learning} based methods is proposed to model spatial and temporal information in different stage. They use CNN to extract Spatial features firstly and then model temporal information by using Long Shot-Term Memory(LSTM) as an encoder to encode the relationship between the sequence-illustrating spatial features. The main problem of these methods is the neglecting of local temporal relationship. 

As a solution to the above problems, our method uses only RGB frames as input and can obtain hybrid features from different level through the multi-scale pyramid attention layer. Moreover, our proposed semantic adversarial learning module can take fully use the video semantic information. which can guide the whole framework to learn more discriminative and semantic video representations.

\section{Proposed Method}
\label{Proposed Method}
In this section, we give detailed description about our method for video classification. Specifically, we first introduce the structure of our method as a whole. Then, we study the multi-scale pyramid attention layer for multiple level features fusion. Finally, we present the semantic adversarial learning module in detail.

\subsection{Overall Architecture}
We design a unified convolutional network that can be divided into three components, \textit{Spatial Network}, \textit{Temporal Network} and \textit{semantic autoencoder}. Figure~\ref{fig: framework} shows the overall architecture of our method. Specifically, we divide a deep convolutional neural network into four \textit{stages}. The output of each stage represents multi-scale features of different visual levels. Then we insert multi-scale pyramid attention (MPA) layer at each stage in order to obtain refined multi-scale features. Given a video clip in the form of $n$ frames sequence $\{\theta_1, \theta_2, \dots, \theta_n\}$, we can obtain four different levels of features $\{V_1^i, V_2^i, \dots, V_n^i\}$, $i \in \{1, 2, 3, 4\}$, after MPA layer of each stage. $V_j^i$ can be rewritten as:
\begin{equation}
V_j^i = \mathrm{Attn}(\theta_j: W_i),
\end{equation}
where $\mathrm{Attn}(\theta_j: W_i)$ is a function representing MPA layer after the $i$-th stage with its parameters $W_i$ operating on the frame $\theta_j$.
The multi-scale features can be formulated as:
\begin{equation}
	\bar{V} = \sum\limits_{i=1}^{4}{\alpha}_i{\cdot}\mathcal{G}(V_1^i, V_2^i, \dots, V_n^i),
\end{equation}
where $\mathcal{G}$ is the frames consensus function, which is able to combine the features from multiple frames to obtain a consensus of representations among them. The $\alpha_i$ is a weight parameter about of $i$-th level, which can be learned automatically.
In order to obtain video semantics, we design \textit{semantic autoencoder}, taking video labels as input. Then semantic adversarial learning module is adopted to guide the architecture to learn semantic representations. We can formulize the adversarial loss as:
\begin{equation}
	\mathcal{L}_{adv}(S, \bar{V}) = \mathrm{Adv}(\bar{V}, S),
	\label{3}
\end{equation}
where $S$ is video semantic extracted from \textit{semantic autoencoder}. After that, the proposed framework can thus generate semantically rich representation $\widetilde{V}$. We note that
$\widetilde{V_k}$ is the representation of the $k$-th video, and the class score $C_k$ is the output of classification layer with $\widetilde{V_k}$ as input.  Combining with standard categorical cross-entropy loss, the final loss function is formed as:
\begin{equation}
	\mathcal{L}_{cls}(y, C) = - \sum\limits_{i=1}^{K}y_i(C_i-\mathrm{log}{\sum\limits_{j=1}^{K}{\mathrm{exp}C_j}}),
	\label{4}
\end{equation}
where $K$ is the number of video classes and $y_i$ is the groundtruth label concerning class $i$. Finally, by combining Eq.~\ref{3} and Eq.~\ref{4}, we can get the final loss function as:
\begin{equation}
	\mathcal{L} = \mathcal{L}_{cls}(y, C) + \mathcal{L}_{adv}(S, \bar{V}).
	\label{5}
\end{equation}

By minimizing $\mathcal{L}$, the objective of video classification can be achieved. Next, we will illustrate the proposed multi-scale pyramid attention layer and semantic adversarial learning module in detail.

\begin{figure*}[!t]
	\centering
	\includegraphics[width=9cm,height=4.6cm]{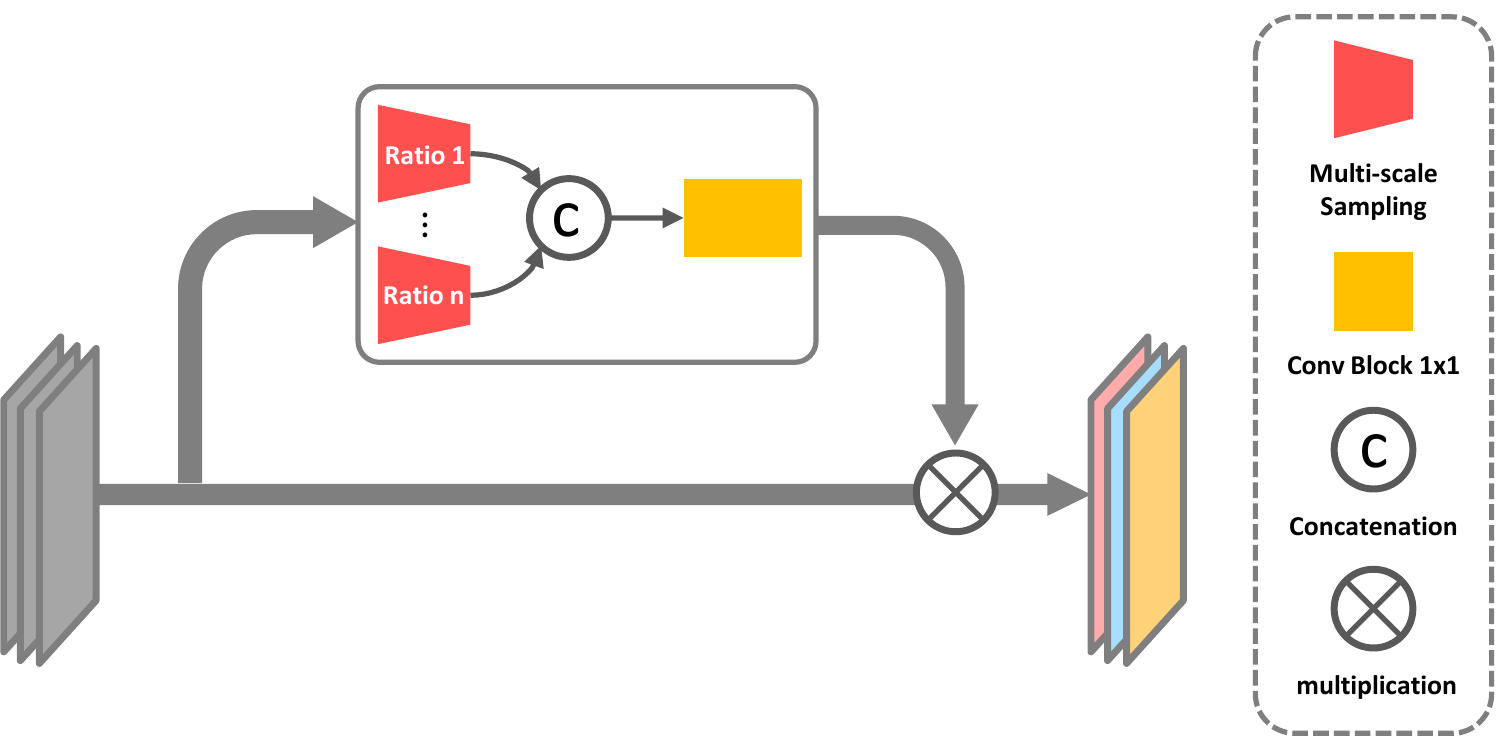}
	\vspace{-0.2cm}
	\caption{Multi-scale pyramid attention unit. We first put feature maps to multi-scale sampling branch to obtain multi-scale information. Then we concatenate them and feed them into a $1\times1$ convolutional layer to merge the information as well as generate the attention weights of feature maps. Finally we obtain the refined video representation.}
	\label{fig: MPA}
\end{figure*}

\subsection{Multi-Scale Pyramid Attention}
The deep convolutional neural network extraction feature is a process from low-level visual features to high-level semantic features. Although the higher network layer is able to extract the global information , it will inevitably lose the details. Therefore, we intend to collect different levels features from a unified convolutional neural network. Specifically, we divide a network (such as ResNet101) into four \textit{stages}, making each halve the resolution of the previous. Each stage contains multiple convolutional layers operating on feature maps of the same resolution. Then we can obtain four sets of features containing high-level semantic information and low level detailed information. However, these multi-scale features are too redundant for the classification task and may degrade the performance. So it is necessary to refine them for classification, and meanwhile, maintaining the multi-scale properties. Motivated by recent progress on residual learning, we introduce a novel multi-scale pyramid attention (MPA) layer that enables the network to consider the importance of each stage feature maps comprehensively with the informations of different receptive fields, so as to obtain reasonable attention weights. The structure of MPA is shown in Figure~\ref{fig: MPA}. It is a pluggable architecture and we put it to the end of each stage, as the Figure~\ref{fig: framework} shows.

Considering the importance of each stage feature maps from $T$ scales, the attention weights $h^i$ of the $i$-th stage feature maps $f^i$ can be formulated as:

\begin{equation}
	h^i = {\tau}(\sum\limits_{t=1}^{T}{\beta_t}l_t(f^i: W_{att}^t)),
\end{equation}
where $\tau$ is the function corresponding to the $1\times1$ convolutional layer. $\beta_t$ is a weight that can be learned automatically. $l_t$ represents the extractor of the $t$-th scale and $W_{att}^t$ is its parameters. Therefore, the function of MPA layer can be rewritten as:
\begin{equation}
	\begin{array}{ll}
		\bar{V}&= \sum\limits_{i=1}^{4}{\alpha}_i{\cdot}\mathcal{G}(V_1^i, V_2^i, \dots, V_n^i) \\
					 &= \sum\limits_{i=1}^{4}{\alpha}_i{\cdot}\mathcal{G}(\mathrm{Attn}({\theta_1}: W_i), \dots, \mathrm{Attn}({\theta_n}: W_i)) \\
					 &= \sum\limits_{i=1}^{4}{\alpha}_i{\cdot}\mathcal{G}(h_1^i{\cdot}f_1^i, \dots, h_n^i{\cdot}f_n^i).
					  
	\end{array}
\end{equation}
Then we can easily obtain the video representation with hybrid multi-scale information.

\subsection{Semantic Adversarial Learning}
Although multi-scale features have the ability to model video information, it still requires the guidance of explicit semantics. So, it is necessary to explore the exact semantic of videos. To this end, the \textit{semantic autoencoder} is introduced in the proposed method. The structure of \textit{semantic autoencoder} contains three fully connected layers supervised by video labels. After training, we freeze trained encoder to generate exact semantic information. Specifically, the semantic $S$ of videos can be written as:
\begin{equation}
	S = \mathrm{encoder}(y),
	\label{8}
\end{equation}
where $y$ is the groundtruth label of videos.  

In order to eliminate the difference of ``real" semantic $S$ and ``fake" semantic $\bar{V}$, we design a semantic adversarial learning module because of its excellent ability of perfectly model the data distribution \cite{goodfellow2014generative,Li_2018_CVPR}. The adversarial loss function is used to encourage $\bar{V}$ close to $S_i$ on the manifold to preserves semantics, by "fooling" a discriminator network $D$ that outputs the probabilities to ensure $\bar{V}$ is as "real" as $S$. The adversarial loss function is formulated as:
\begin{equation}
	\begin{array}{ll}
		\mathcal{L}_{adv}(S, \bar{V})&= \mathrm{Adv}(\bar{V}, S) \\
									 &= \mathbb{E}_S(\mathrm{log}D(S)) + \mathbb{E}_{\bar{V}}(\mathrm{log}[1-D(\bar{V})]),
	\end{array}
\end{equation}
where $\bar{V}$ can be regarded as a transformation $G$ of frames sequence $\theta$. So the loss function can be rewritten as:
\begin{equation}
	\mathcal{L}_{adv}(S, \bar{V})=\mathbb{E}_S(\mathrm{log}D(S)) + \mathbb{E}_{\theta}(\mathrm{log}[1-D(G(\theta))]),
\end{equation}
where $G$ tries to minimize $\mathcal{L}_{adv}$ against $D$ that tries to maximize it, \textit{i.e.,} $D^* = {\mathrm{argmin}_G{\mathrm{max}_D}{\mathcal{L}_{adv}(S, \bar{V})}}$ For better gradient of learning $G$, we actually minimize $\mathbb{E}_{\theta}(\mathrm{log}[-D(G(\theta))])$ instead of $\mathbb{E}_{\theta}(\mathrm{log}[1-D(G(\theta))])$. Therefore, the final adversarial loss function is defined as:
\begin{equation}
	\mathcal{L}_{adv}(S, \bar{V})=\mathbb{E}_S(\mathrm{log}D(S)) + \mathbb{E}_{\theta}(\mathrm{log}[-D(G(\theta))]),
\end{equation}
Combining with Eq.~\eqref{5}, we can obtain the optimization of our proposed method. For \textit{Spatial Network}, the whole network can be divided into three parts: semantic generator G, classification layer C and semantic discriminator D. We adopt an alternating optimization to train all three parts mentioned above to avoid gradient vanishing problem caused by the minmax loss. 

Firstly, semantic discriminator D is trained by minimizing Eq.~\eqref{5}. Update the parameters of D with G and C fixed:
\begin{equation}
	D^* = \mathrm{arg}{\min\limits_{D}}\mathcal{L},
\end{equation}
Then, semantic generator G and classification layer C are trained by minimizing Eq.~\eqref{5}. Update the parameters of G and C with D fixed:
\begin{equation}
G^*, C^* = \mathrm{arg}{\min\limits_{G, C}}\mathcal{L}.
\end{equation}
The whole semantic adversarial learning module (SAL) is summarized in Algorithm 1. The \textit{Temporal Network} has the same setting as \textit{Spatial Network}.

\begin{algorithm}[!t]
	\renewcommand{\algorithmicrequire}{ \textbf{Input:}}
	\renewcommand{\algorithmicensure}{ \textbf{Output:}}
	\caption{The optimization algorithm of SAL}
	\begin{algorithmic}[1]
		\REQUIRE   video dataset $\mathcal{V} = \{({\theta}_k, y_k), k=1,\dots,K\}$ \\
		\ENSURE    semantic representation $\bar{V}$, class score $C$ 
		\STATE Initialized D, G, C; pretrained senmantic encoder
		\REPEAT
		\STATE Extract $S$ by Eq.~\ref{8}.
		\STATE Fixing parameters of G and D
		\STATE Update D by minimizing Eq.~\ref{5}. ($S \to 1$, $\bar{V} \to 0$)
		\STATE Fixing parameters of G and D
		\STATE Update G, C by minimizing Eq.~\ref{5}. ($\bar{V} \to 1$)
		\STATE Extract $\bar{V}$ and $C$ 
		\UNTIL convergence
	\end{algorithmic}
\end{algorithm}

\section{Experimental Evaluations}
\label{Experiments Evaluations}

In this section, evaluation datasets and implementation details used in experiments will be first introduced. Then we will study different aspects of our proposed modal to verify the effectiveness, respectively. Finally, we will make a comparison between our model with other RGB based state-of-the-art methods and provide a visualization of our experimental results.

\subsection{Datasets and Implementation Details}
\textbf{Evaluation Datasets.} In order to evaluate our proposed model, we conduct action recognition experiments on two popular video benchmark datasets: UCF101 \cite{soomro2012ucf101} and HMDB51 \cite{kuehne2011hmdb}. The UCF101 dataset are collected from the Internet, containing 13,320 videos which are divided into 101 classes. While the HMDB51 dataset are collected from the realistic videos, including movies and web videos, containing 6,766 videos which are divided into 51 action categories. We follow the officially offered scheme which divides dataset into 3 training and testing splits and finally report the average accuracy over the three splits. For \textit{Spatial network}, we directly utilize RGB frames extracted from videos. For \textit{Temporal Network}, the difference between adjacent frames is used to model temporal information of videos.

\noindent\textbf{Implementation details.} In generation procedure, we use stochastic gradient descent algorithm to train our \textit{Spatial Network} and \textit{Temporal Network}. The mini-batch size is set to 64 and the momentum is set to 0.9. The initial learning rate is set to 0.001 for \textit{Spatial Network} and \textit{Temporal Network} and decreases by 0.1 every 40 epochs. 

For adversarial training procedure, we use adaptive moment estimation algorithm to train D Network and the initial learning rate is set to 0.0001. The training procedure of \textit{Spatial Network} and \textit{Temporal Network} stops after 80 epochs and 120 epochs respectively. We use gradient clipping of 20 and 40 for Spatial and Temporal training procedure to avoid gradient explosion. We train our model with 4 NVIDIA TITAN X GPUs and all the experiments are implemented under the Pytorch.

\begin{table}[!t]
	\begin{center}
		\caption{Performance comparison on HMDB51 split 1.}
		\vspace{0.15cm}
		\renewcommand\arraystretch{1.5}
		\begin{tabular}{|p{2cm}<{\centering}|p{1.5cm}<{\centering}|p{1.5cm}<{\centering}|p{1.5cm}<{\centering}|}
			\hline			
			\multirow{1.2}{*}{Method} & \multirow{1.2}{*}{Spatial} &\multirow{1.2}{*}{Temporal} & \multirow{1.2}{*}{Combine} \\			
			\hline
			\multirow{1.3}{*}{TSN}
			& 52.0\%  & 57.1\% & 62.5\%  \\	
			\hline
			\multirow{1.3}{*}{\textbf{Ours}}
			& \textbf{54.4\%}  & \textbf{57.8\%} & \textbf{63.7}\% \\			
			\hline		
		\end{tabular}
	\end{center}
\end{table}
\begin{table}[!t]
	\begin{center}
		\caption{Performance comparison on UCF101 split 1.}
		\vspace{0.15cm}
		\renewcommand\arraystretch{1.5}
		\begin{tabular}{|p{2cm}<{\centering}|p{1.5cm}<{\centering}|p{1.5cm}<{\centering}|p{1.5cm}<{\centering}|}
			\hline			
			\multirow{1.2}{*}{Method} & \multirow{1.2}{*}{Spatial} &\multirow{1.2}{*}{Temporal} & \multirow{1.2}{*}{Combine} \\			
			\hline
			\multirow{1.3}{*}{TSN}
			& 84.5\%  & 85.4\% & 89.9\%  \\		
			\hline
			\multirow{1.3}{*}{\textbf{Ours}}
			& \textbf{86.1}\%  & \textbf{86.8}\% & \textbf{91.2}\% \\			
			\hline
		\end{tabular}
	\end{center}
\end{table}
\begin{table}[!t]
	\begin{center}
		\caption{Ablation study of our proposed method.}
		\vspace{0.2cm}
		\renewcommand\arraystretch{1.5}
		\begin{tabular}{|p{1.5cm}<{\centering}|p{2.6cm}<{\centering}|p{1.2cm}<{\centering}|p{1.2cm}<{\centering}|}
			\hline	
			\multirow{1.2}{*}{Dataset} & \multirow{1.2}{*}{Method} &\multirow{1.2}{*}{Network} & \multirow{1.2}{*}{Accuracy} \\	
			\hline	
			\multirow{3.5}{*}{HMDB51}
			& Baseline & Spatial & 52.0\%  \\
			\cline{2-4}
			& MPA & Spatial & 54.0\%   \\
			\cline{2-4}
			& MPA + SAL  & Spatial &\textbf{54.4\%} \\
			\hline
			\multirow{3.5}{*}{UCF101}
			& Baseline & Spatial & 84.5\%  \\
			\cline{2-4}
			& MPA & Spatial & 85.2\%  \\
			\cline{2-4}
			& MPA + SAL   & Spatial &\textbf{86.1\%} \\		
			\hline
		\end{tabular}
	\end{center}
\end{table}
\begin{table*}[!t]
	\begin{center}
		\caption{Comparison with state-of-the-art methods on the UCF101 and HMDB51 datasets. The accuracy is reported as average over three splits. For fair comparison, we consider methods that use only RGB input. It can be seen our method obtains the best performance.}
		\vspace{0.3cm}
		\renewcommand\arraystretch{1.5}
		\begin{tabular}{|p{10cm}<{\centering}|p{3.5cm}<{\centering}|p{1.2cm}<{\centering}|p{1.2cm}<{\centering}|}
			\hline			
			\multirow{1.2}{*}{Method} & \multirow{1.2}{*}{Pre-train} &\multirow{1.2}{*}{UCF101} & \multirow{1.2}{*}{HMDB51} \\			
			\hline
			HOG\cite{wang2013action} & None  & 72.4\% & 40.2\%  \\
			\hline
			ConvNet+LSTM\cite{donahue2015long} & \tabincell{c}{ImageNet}  & 68.2\% & -  \\
			\hline
			Two Stream Spatial Network\cite{simonyan2014two}  & ImageNet  & 73.0\% & 40.5\%  \\
			\hline
			Conv Pooling Spatial Network\cite{feichtenhofer2016convolutional}  & ImageNet  & 82.6\% & -  \\
			\hline
			Spatial Stream ResNet & ImageNet  & 82.3\% & 43.4\%  \\
			\hline
			Spatial TDD\cite{wang2015action} & ImageNet  & 82.8\% & 50.0\%  \\
			\hline
			TSN Spatial Network\cite{wang2016temporal} & ImageNet  & 86.4\% & 53.7\%  \\
			\hline
			TSN (RGB+RGB-Diff)\cite{wang2016temporal} & ImageNet  & 91.0\% & -  \\
			\hline
			RGB-I3D\cite{carreira2017quo} & ImageNet  & 84.5\% & 49.8\%  \\
			\hline
			CoViAR\cite{wu2018compressed} & ImageNet & 90.4\% &59.1\% \\
			\hline
			DCD\cite{zhao2018recognize}  & ImageNet  & 91.8\% & -  \\
			\hline
			LTC\cite{varol2018long}  & Sports-1M & 82.4\% & 48.7\%  \\
			\hline
			C3D\cite{tran2015learning} & Sports-1M & 85.8\% & 54.9\%  \\
			\hline
			Pseudo-3D Resnet\cite{qiu2017learning}  & ImageNet+Sports-1M & 88.6\% & - \\
			\hline
			C3D\cite{tran2015learning} & Kinetics & 89.8\% & 62.1\%  \\
			\hline
			\textbf{Ours}& ImageNet  & \textbf{92.7\%} & \textbf{66.3\%} \\			
			\hline

		\end{tabular}
	\end{center}
\end{table*}

\subsection{Results and Ablation Study}
In this subsection, we will investigate the performance of our proposed method. The analysis for the performance of single and multiple modalities. All the results are trained with the same network backbone and strategies illustrated in previous sections for fair comparison. 

We first evaluate the effectiveness of our proposed method. In this section, we compare the performance between ours and TSN \cite{wang2016temporal} with the same experimental condition. The experiment is performed on HMDB51 split 1 and UCF101 split 1, the results are summarized in Table 1 and Table 2. As is shown in them, our method is superior than TSN both in spatial and temporal branch. In spatial branch, our method has about 2\% improvement in performance compared to TSN, which proves the effectiveness of proposed MPA and SAL. but in temporal branch, it is a limited improvement because differential image has less visual elements than RGB.  

In order to justify the effectiveness of our proposed MPA and SAL, we conduct an ablation study for them. All experiments in this ablation study are performed on the split 1 of HMDB51 and UCF101 by \textit{Spatial Network}. The results are shown in Table 3. The \textit{baseline} means the original network without multi-scale pyramid attention and semantic adversarial learning module. The \textit{MPA} means add multi-scale pyramid attention layer to \textit{baseline}. It has about 1\%-2\% improvement in performance compared to \textit{baseline}, which can prove the effectiveness of proposed MPA. The \textit{MPA+SAL} means add both multi-scale pyramid attention and semantic adversarial learning module. It has about 1\% improvement in performance compared to \textit{MPA}, that can prove the effectiveness of the proposed SAL.

\begin{figure*}[!t]
	\centering
	\includegraphics[width=16cm,height=7.5cm]{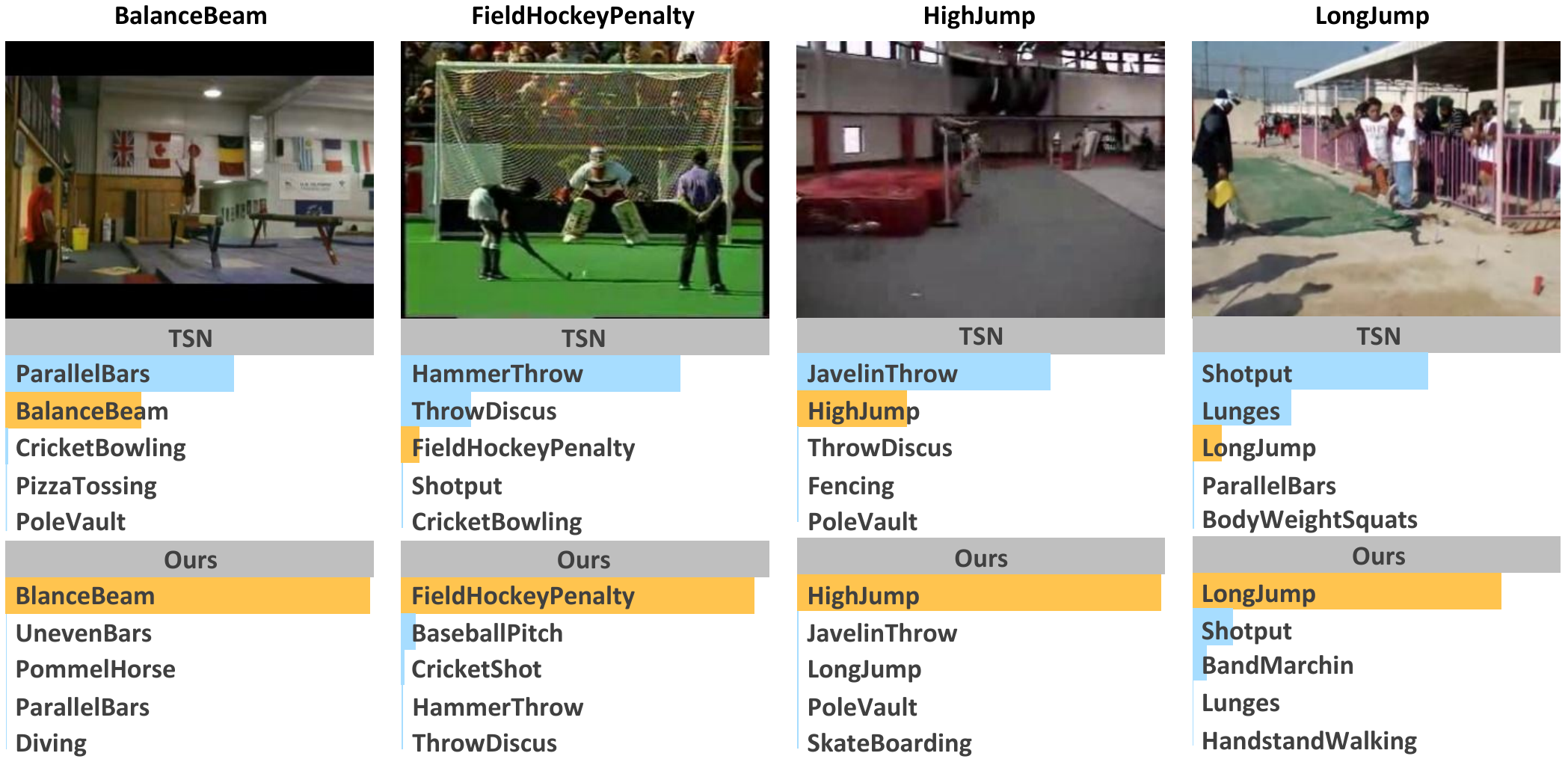}
	\vspace{-0.2cm}
	\caption{A comparison of top-5 predictions between TSN and our proposed method on UCF101 split 1. The tags on the top is the groundtruth labels. The yellow bars indicate correct classifications and the blue stand for incorrect cases. The length of each bar shows its confidence.}
	\label{fig: acc_visualization}
	\vspace{-0.25cm}
\end{figure*}
\begin{figure}[!t]
	\centering
	\includegraphics[width=7cm,height=6cm]{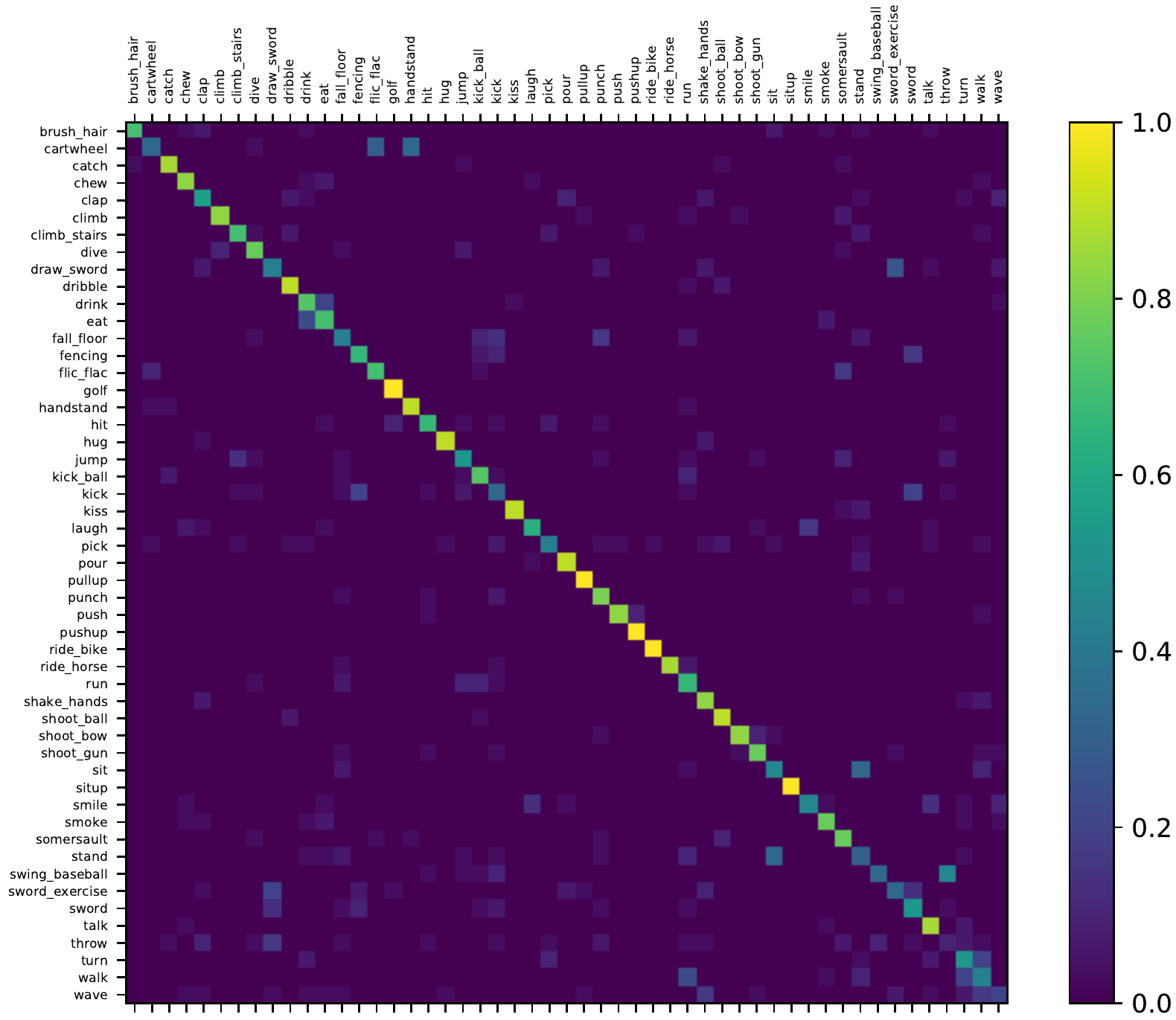}
	\vspace{-0.3cm}
	\caption{Confusion matrix for HMDB51 dataset.}
	\label{fig: hmdb51_cf}
\end{figure}
\begin{figure}[!t]
	\centering
	\includegraphics[width=7cm,height=6cm]{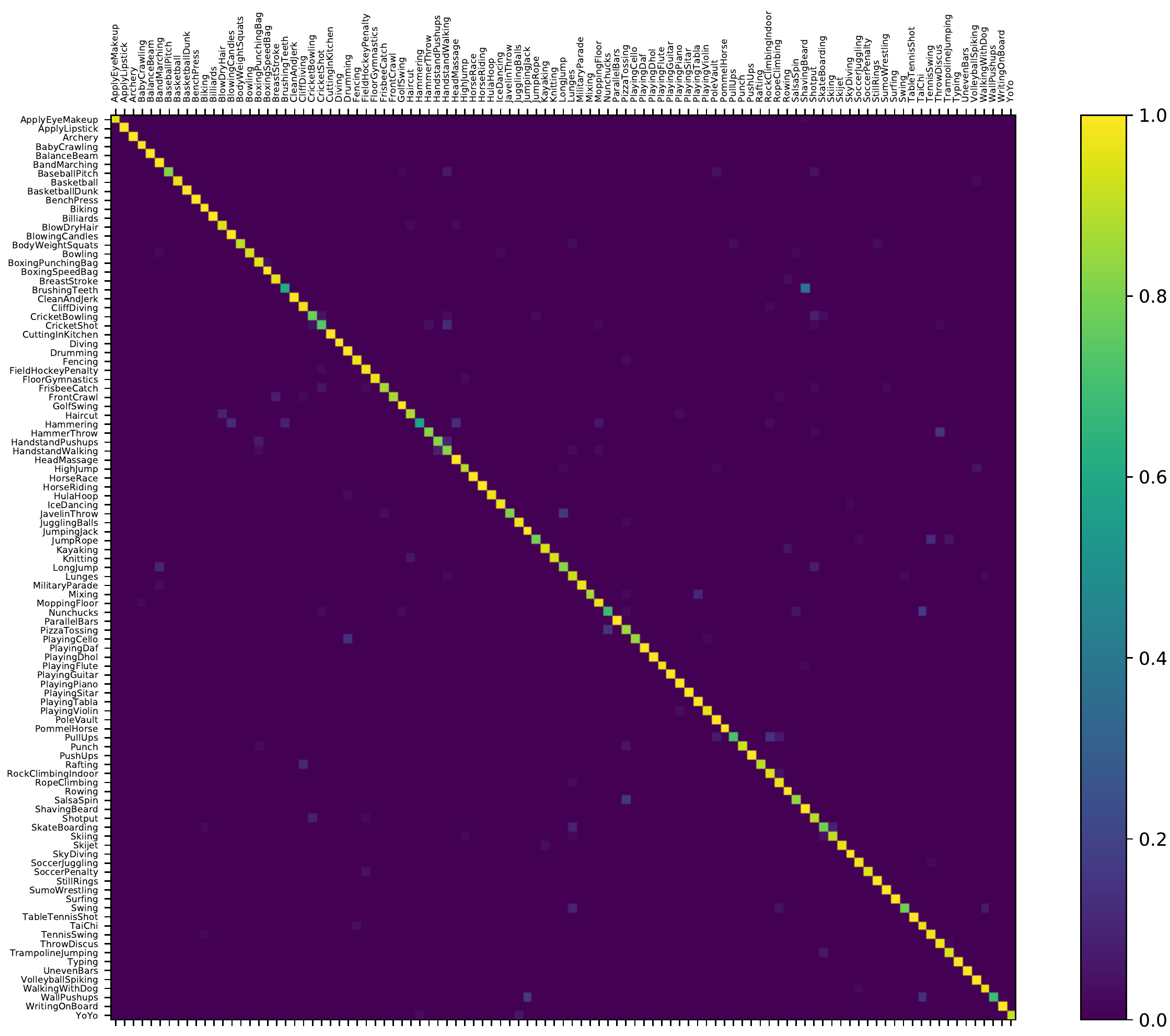}
	\vspace{-0.3cm}
	\caption{Confusion matrix for UCF101 dataset.}
	\label{fig: ucf101_cf}
\end{figure}

\subsection{Comparison with the State-of-the-Arts}
In this subsection we compare the classification performance of our approach with other state-of-the-art methods that take RGB frames as input. The experiment is conducted on two popular video action recognition benchmarks: UCF101 and HMDB51. The results are shown in Table 4, where we compare our method with traditional approach HOG and a series of deep learning based methods such as 3D convolutional network, trajectory-pooled deep convolutional descriptors, temporal segment network and compressed video action recognition. It can be seen that our model achieves best results than other methods on these benchmark datasets, which can demonstrates the advantage of our proposed method and the effectiveness of multi-scale feature semantic modeling.

\subsection{Visualization}
In this subsection, we present a qualitative classification results. Figure~\ref{fig: acc_visualization} illustrate the comparison of top-5 predictions between TSN and our proposed method on UCF101 split 1. The results show that the original two-stream based methods (such as TSN) are easily fooled by common background. For instance, it regards BalanceBeam as ParallelBars, since the similar backgroud of gym and equipments. The reasons is that those methods fail to capture details and local context information effectively. The results base on our method can classify this actions will, which demonstrate the effectiveness of proposed MPA and SAL module. Figure~\ref{fig: hmdb51_cf} and Figure~\ref{fig: ucf101_cf} show the confusion matrix for HMDB51 and UCF101, respectively. All the results can demonstrate the superior performance of our method.

\section{Conclusion}
\label{sec: Conclusions}
In this paper, we proposed a new deep network architecture for video classification. The proposed architecture can only need RGB frames as input to model spatial information and temporal information for videos, which greatly reduce the computational cost compared with those methods using optical flow. In order to obtain more discriminative video representation, we design multi-scale pyramid attention (MPA) to refine and merge different level features. Then we introduce semantic adversarial learning (SAL) module to guide learning procedure and to generate more discriminative semantic representation. Comprehensive experiment results on two popular benchmark datasets show that our method yields state-of-the-art performance in video classification tasks.

\section{Acknowledgments}
Our work was supported in part by the National Natural Science Foundation of China under Grant 61572388 and 61703327, in part by the Key R\&D Program-The Key Industry Innovation Chain of Shaanxi under Grant 2017ZDCXL-GY-05-04-02, 2017ZDCXL-GY-05-02 and 2018ZDXM-GY-176, and in part by the National Key R\&D Program of China under Grant 2017YFE0104100.

\bigskip
\bibliographystyle{aaai}
\bibliography{AAAI-XieD.2866}

\end{document}